\documentclass{article}

\usepackage{arxiv}
\usepackage[utf8]{inputenc} 
\usepackage[T1]{fontenc}    
\usepackage{hyperref}       
\usepackage{url}            
\usepackage{booktabs}       
\usepackage{amsfonts}       
\usepackage{nicefrac}       
\usepackage{microtype}      
\usepackage{lipsum}		
\usepackage{graphicx}
\usepackage{natbib}
\usepackage{doi}
\usepackage{wrapfig}
\usepackage{algorithm}
\usepackage[noend]{algpseudocode}
\usepackage[bottom]{footmisc}
\usepackage{amsmath}
\usepackage{xcolor}         
\newtheorem{theorem}{Theorem}
\newtheorem{lemma}{Lemma}

\newtheorem{proposition}{Proposition}

\newenvironment{proof}{\paragraph{Proof.}}{\hfill$\square$}

\title{On Leaky-Integrate-and Fire as Spike-Train-Quantization Operator on Dirac-Superimposed Continuous-Time Signals}

\author{ 
{
\hspace{1mm}Bernhard A.~Moser}\thanks{double affiliation: Software Competence Center Hagenberg (SCCH), 4232 Hagenberg, Austria} \\
	Institute of Signal Processing \\
	Johannes Kepler University of Linz\\
	\texttt{bernhard.moser@\{scch.at,jku.at\}} 
	\And
	{\hspace{1mm}Michael Lunglmayr} \\
	Institute of Signal Processing\\
	Johannes Kepler University of Linz, Austria\\
	\texttt{michael.lunglmayr@jku.at} 
	}

\renewcommand{\shorttitle}{\textit{arXiv} Signal Space for Leaky Integrate-and-Fire}

\hypersetup{
pdftitle={A template for the arxiv style},
pdfsubject={q-bio.NC, q-bio.QM},
pdfauthor={Bernhard A.~Moser, Michael Lunglmayr},
pdfkeywords={Leaky Integrate-and-Fire, Neuromorphic Computing, Spiking Neuron Model, Threshold-based Sampling},
}

\begin{document}
\maketitle

\begin{abstract}
Leaky-integrate-and-fire (LIF) is studied as a non-linear operator that maps an integrable signal $f$ to a sequence $\eta_f$ of discrete events, the spikes.
In the case without any Dirac pulses in the input, it makes no difference whether to set the neuron's potential to zero or to subtract the 
threshold $\vartheta$ immediately after a spike triggering event. However, in the case of superimpose Dirac pulses  the situation is different which raises the question of a mathematical justification of each of the proposed reset variants. 
In the limit case of zero refractory time the standard reset scheme based on threshold subtraction results in a modulo-based reset scheme which 
allows to characterize LIF as a quantization operator based on a weighted Alexiewicz norm $\|.\|_{A, \alpha}$ with leaky parameter $\alpha$.
We prove the  quantization formula $\|\eta_f - f\|_{A, \alpha} < \vartheta$ under the general condition of local integrability, almost everywhere boundedness  and locally finitely many superimposed weighted Dirac pulses which provides a much larger signal space and more flexible sparse signal representation than manageable by classical signal processing.
\end{abstract}


\section{Introduction}
\begin{figure}
	\centering
		\includegraphics[width=0.5\textwidth]{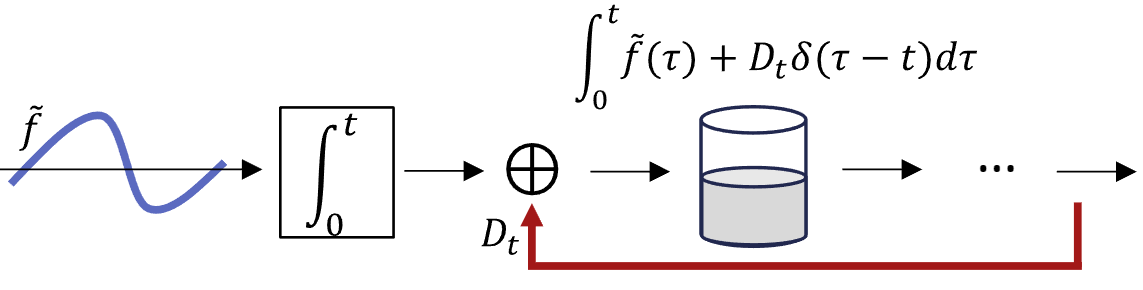}		
  		\caption{Schematic description of integration within a LIF neuron with spike feedback connection leading to signals with superimposed Dirac pulses.		}
	\label{fig:Dirac}
\end{figure}

In contrast to more bio-physically realistic models such as the Hodgkin-Huxley model, LIF is a middle ground that captures essential features of bio-inspired time-based information processing and is simple enough to be applicable from a neuromorphic engineering and signal processing perspective~\cite{bookGerstner2014,Nunes2022}. In biology in particular, spikes usually have different forms, and their generation and re-initialization follow complex dynamics, 
as shown by the Hodgkin-Huxley differential equations. In this paper, we look at LIF from a mathematical point of view by sticking to its characterizing idealization principles. (1),  shape information is neglected by idealizing spikes as shapeless weighted Dirac pulses, (2), the spike triggering process is realized by means of thresholding instead of utilizing differential equations, and, (3), the process of triggering and re-initialization is idealized by instantaneously acting events. A central mathematical question is about the information being preserved by applying LIF on an input signal and how it can be quantified. 
Although not ubiquitous in nature, signals being continuous in time  and superimposed by Dirac pulses are typically encountered in a perception task based on recurrent spiking neural networks. Consider the situation that a LIF neuron is used for data acquisition from an analogue input signal then there might occur an additional superimposed input spike coming from a feedback connection from within the network.
Fig.~\ref{fig:Dirac} graphically depicts the LIF integration in such a scenario. 
The integration is schematically described as an addition of the function value to a previously integrated level. The feedback spikes are typically added directly to the level in an implementation (red arrow), hence causing discontinuous jumps of the neuron's potential. As the potential is modeled by an integral, the jumps become equivalent to corresponding superimposed Dirac pulses on the integrand.


Our approach is inspired by the quasi-isometry relation for threshold-based sampling in a more general setting of Weyl's discrepancy measure, 
see~\cite{Moser2017Similarity,MoserLunglmayr2019QuasiIsometry}. 
For this paper we rely on the mathematical notions of~\cite{moser2023arXiv_SNNAlexTop}, by referring to $\mbox{LIF}_{\alpha, \vartheta}$ as the operator that maps the signal $f: [0, \infty) \rightarrow \mathbb{R}$ to a spike train by applying LIF with threshold $\vartheta$ and leaky parameter $\alpha$, and by referring to $\|\|_{A, \alpha}$
as the weighted Alexiewicz norm based on $\alpha$, i.e.,
\begin{equation}
\label{eq: AlexNorm}
\|f\|_{A, \alpha}:= \sup_{T}\left|\int_{0}^T e^{-\alpha\, (T- t)} f(t) dt\right| 
\end{equation}
We take up the main result of~\cite{moser2023arXiv_SNNAlexTop}, i.e.,
\begin{equation}
\label{eq:qFormula}
\|\mbox{LIF}_{\alpha, \vartheta}(f) - f\|_{A, \alpha} < \vartheta
\end{equation}
which has been proven for bounded piecewise continuous signals.  

In this paper we will show that (\ref{eq:qFormula}) also holds valid under the general condition on the signals 
$f:[0, \infty) \rightarrow \mathbb{R}$ in terms of 
\begin{itemize}
\item[C1)] {\it local integrability} (w.r.t Lebesgue, Riemann or Henstock-Kurzweil) and bounded Alexiewicz norm, i.e.,  $\|f\|_{A, \alpha}<\infty$,
\item[C2)]  {\it almost everywhere (a.e.) boundedness},  
\item[C3)] {\it locally finitely many (weighted) Dirac impulses}, i.e.,
in any time interval $I=[t_a, t_b]$ there are at most finitely many (weighted) Dirac impulses, and
\item[C4)] $\lim_{\varepsilon \rightarrow 0} |\int_0^{\varepsilon}| = 0$ as initialization condition. 
\end{itemize}

In the signal space condition C1 there is a subtle aspect regarding the conception of the integral due to the fact that integration itself is a matter of mathematical idealization and conception. Usually the Riemann or the Lebesgue integral is taken as the natural choice in a classical signal processing setting, e.g.,
when considering Hilbert spaces and the $L2$-norm. However, in our setting the Alexiewicz norm turns out to play the central role. 
This way it is natural to consider the space of signals and that definition of an integral on which the Alexiewicz norm is well-defined and bounded. 
This leads to the Henstock-Kurzweil (HK) integral as an alternative integral that is even more flexible than its classical counterparts as it can be shown that any Riemann, or, resp. Lebesgue, integrable function is also HK integrable, see~\cite{Bartle2001, KurtzSwartz2004}. In fact, any HK integrable function is also Lebesgue measurable, and $f$ is Lebesgue integrable if and only if both $f$ and $|f|$ are HK integrable.  In condition C2 the signal $f$ is required to be a.e. bounded, i.e.,  $f$ is bounded on $[0,\infty) \backslash U$ where the exceptional set $U$ of potentially unbounded values has Lebesgue measure zero, i.e., $\lambda(U)=0$. In this setting the upper bound is replaced by the essential supremum given by $\mbox{ess-sup} f := \inf \{ U_{f}^{\mbox {\tiny ess}} \}$ where ${U_{f}^{\mbox {\tiny ess} }=\{a\in \mathbb {R} :\lambda (f^{-1}(a,\infty ))=0\}}$, $f^{-1}(M):= \{t: f(t)\in M\}$. Note that $\lambda(\{t \in [a,b]: |f(t)| \leq \mbox{ess-sup} f\}) = b-a$. Condition C3 covers the scope of spike trains (of graded spikes) modeled by a superposition of weighted Dirac impulses with spikes being well-separable in time.
C4 is a technical initialization condition which allows to set $s_0=0$. In the above mentioned scenario of a LIF neuron in a recurrent network and incoming bounded signals, C4 is naturally justified as spikes that are fed back from within a recurrent network can only be triggered for $t > 0$. 

The scope of this signal space is made up by Dirac superimposed signals in continuous time that extends that of~\cite{moser2023arXiv_SNNAlexTop}.
It includes discrete events in terms of spike trains as well as real-valued signal values with potentially unlimited range, as long as they are integrable on any local time interval. 

The paper is outlined as follows. After recalling the LIF model in Section~\ref{s:LIF}, we prove its well-defined behavior under the conditions of
C1-4. In Section~\ref{s:Alex} we generalize the quantization formula of~\cite{moser2023arXiv_SNNAlexTop} to this general signal class. 
Examples demonstrate the different behavior when utilizing different reset regimes of the LIF neuron. 
Finally in Section~\ref{s:Evaluations} we evaluate the found quantization formula on randomly generated spike trains showing 
characteristics depending on the leaky parameter and the selected reset regime.

\section{Leaky-Integrate-and-Fire Revisited}
\label{s:LIF}
Since leaky-integrate-and-fire neuron model (LIF) with leaky  parameter $\alpha>0$ and threshold $\vartheta>0$ is made up by an integration process, 
we need at least the property of {\it local integrability}. Another assumption is needed to guarantee that the sampling process does 
not get stuck in an accumulation point. As shown below for this a necessary and sufficient condition is {\it boundedness almost everywhere}, that is bounded except on a union of time intervals of Lebesgue measure zero.

For this let us sample a {\it locally HK-integrable} and {\it a.e. bounded} signal $f$ with locally finitely many Dirac impulses 
by means of LIF with leaky parameter $\alpha>0$ and threshold $\vartheta>0$, to obtain the spike train $\eta(t) = \sum_k s_k\delta(t - t_k)$, where $s_k \in \mathbb{R}$ denotes the amplitude of the spike at time $t_k$. 
The time points $t_{k}$ are recursively given by
\begin{equation}
\label{eq:LIFsample}
t_{k+1} := \inf\left\{T\geq t_k + t_r: \left|\int_{t_k}^{T} e^{-\alpha (t_{k+1} -  t)} \big(f(t) + r_k \delta(t-t_k)\big) dt\right| \geq \vartheta\right\},
\end{equation}
where $t_r\geq 0$ is the refractory time and $T=t_{k+1}$ is the first time point after $t_{k}$ that causes the integral in 
(\ref{eq:LIFsample}) to violate the sub-threshold condition $|\int_{t_k}^{T} e^{-\alpha (t_{k+1} -  t)} f(t) dt | < \vartheta$.
The term $r_k \delta(t-t_k)$ refers to the reset of the membrane potential in the moment a spike has been triggered.
In the standard definition of LIF for discrete spike trains, see~\cite{bookGerstner2014}, the reset is defined as the membrane potential that results from subtracting the threshold if the membrane's potential reaches the positive threshold level $+\vartheta$, respectively adding $\vartheta$ if 
the membrane's potential reaches the negative threshold level $-\vartheta$. 
In the case of bounded $f$ the integral $g(t):= \int_{t_k}^t e^{-\alpha (t_{k+1} -  t)} f(t) dt$ is changing continuously in $t$ 
so that the threshold level in (\ref{eq:LIFsample}) is exactly hit. 
Consequently the resulting reset amounts to zero, i.e., $r_k = 0$ and the resulting amplitude $s_k$ of the triggered spike is defined accordingly, i.e., $s_k = +\vartheta$, when the positive threshold value is reached, and $s_k = -\vartheta$ when the negative threshold value is reached.
For a mathematical analysis and a discussion of how to define the reset $r_k$  in the presence of Dirac impulses see~\cite{moser2023arXiv_SNNAlexTop}.

\begin{figure}
	\centering
		\includegraphics[width=0.325\textwidth]{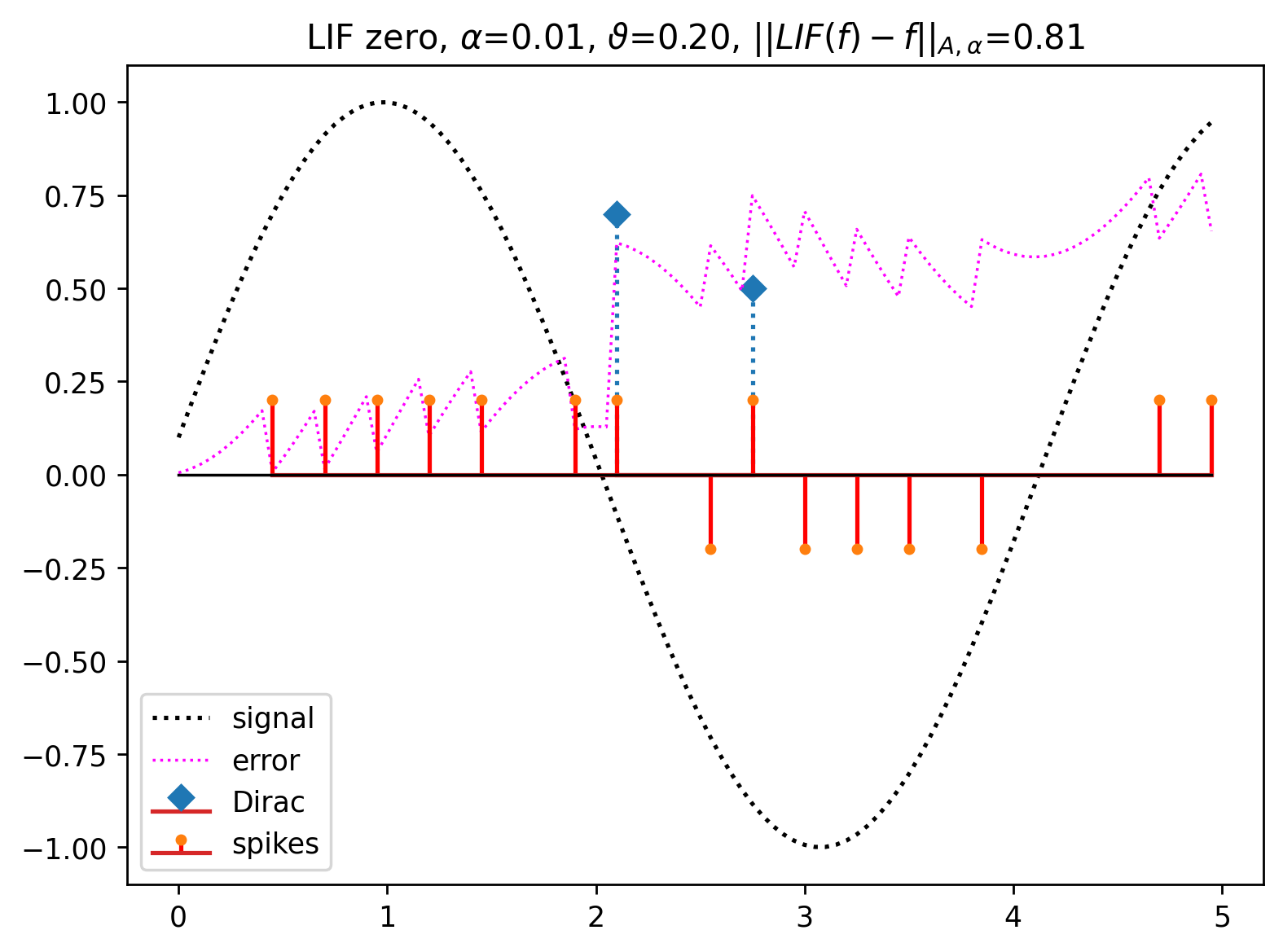}
		\includegraphics[width=0.325\textwidth]{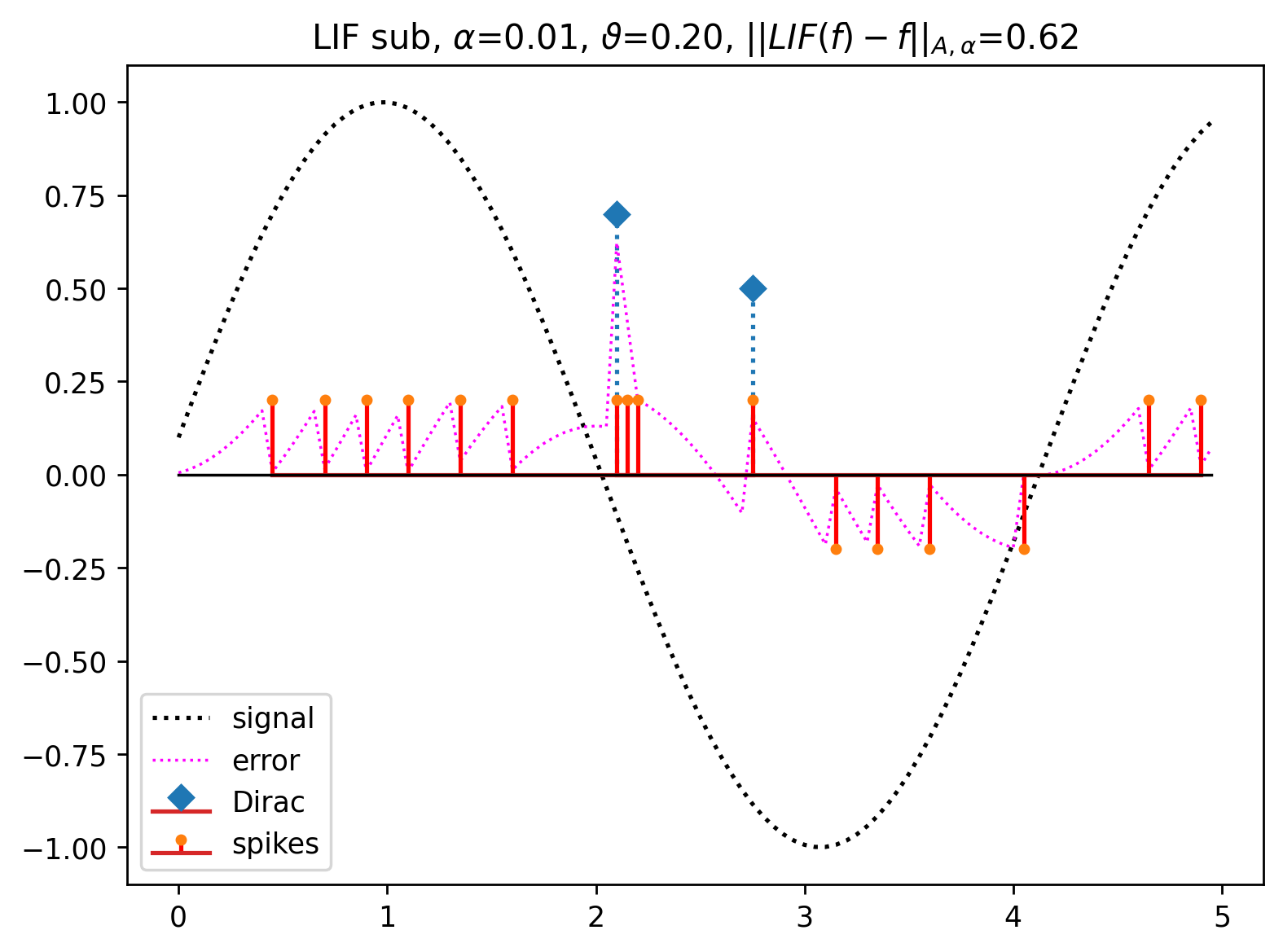}
		\includegraphics[width=0.325\textwidth]{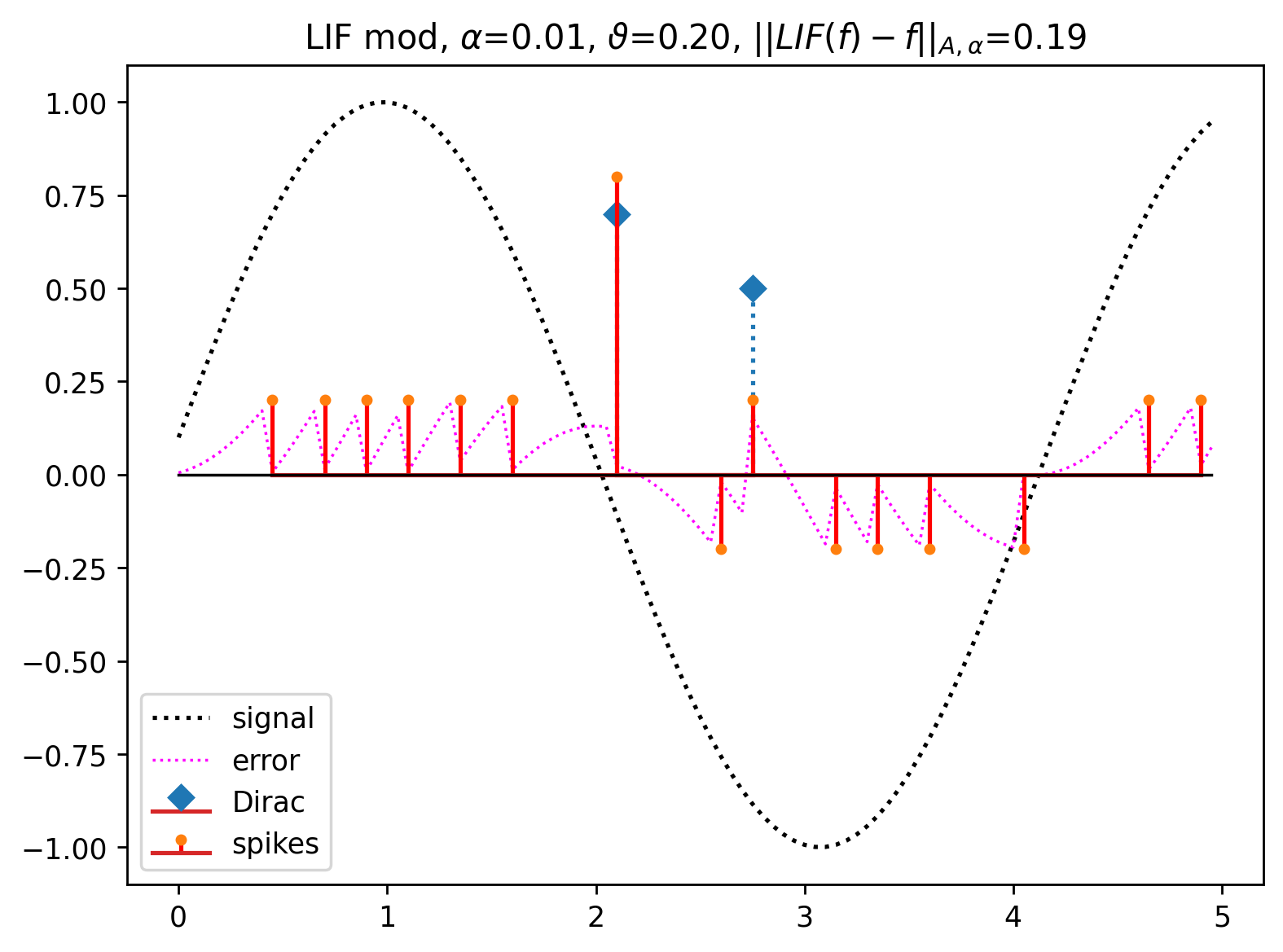}
		\includegraphics[width=0.325\textwidth]{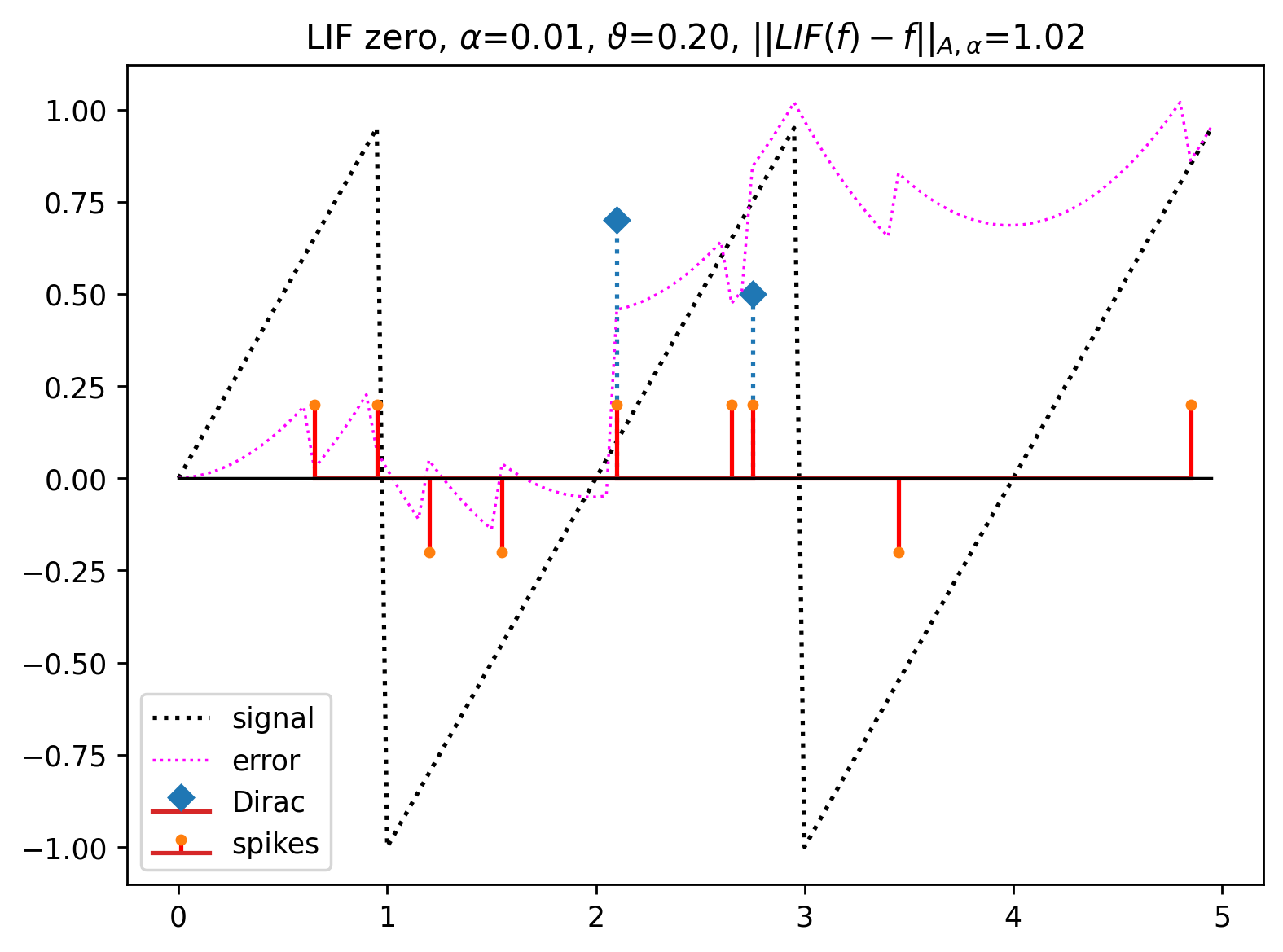}
		\includegraphics[width=0.325\textwidth]{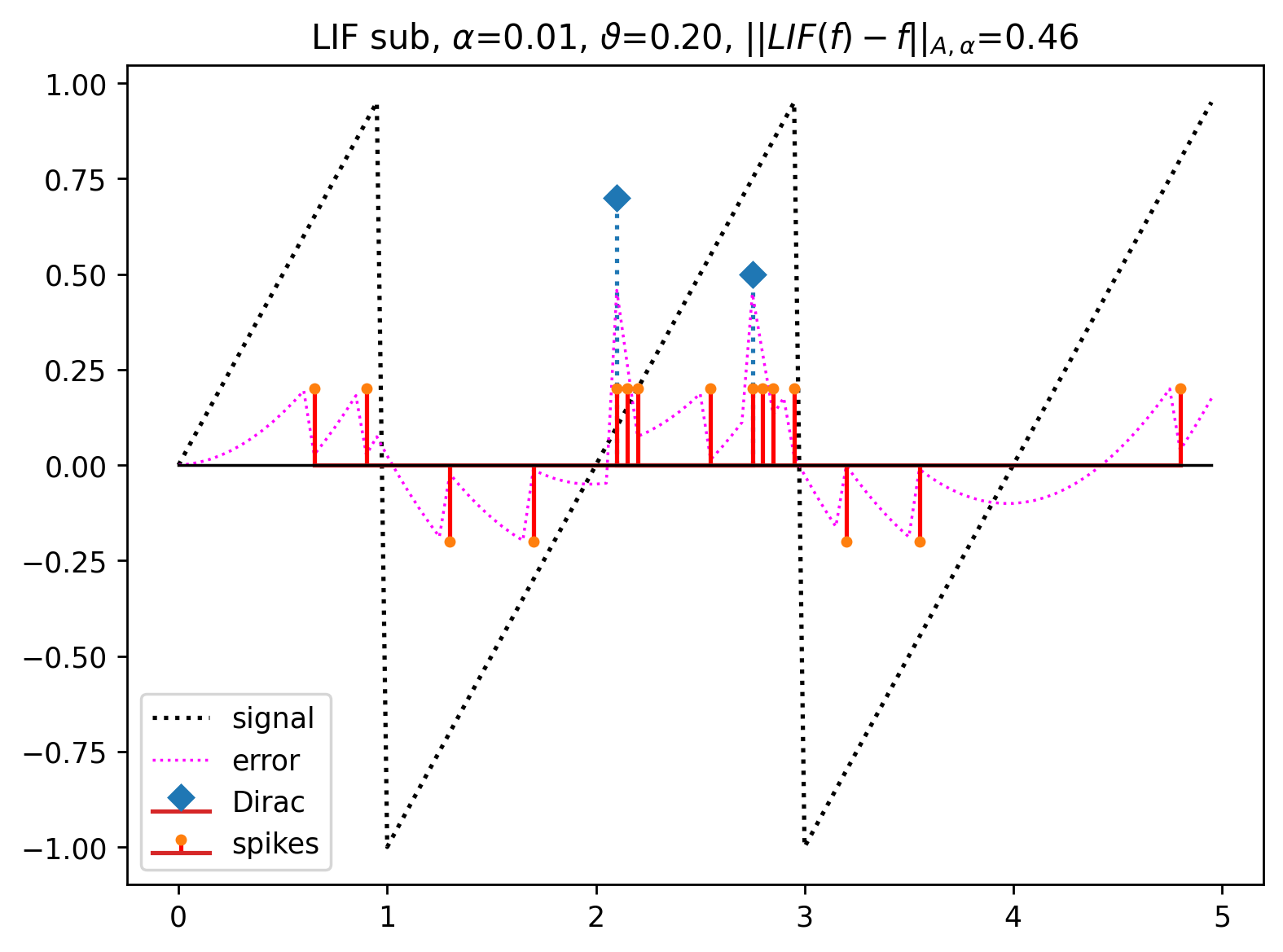}
		\includegraphics[width=0.325\textwidth]{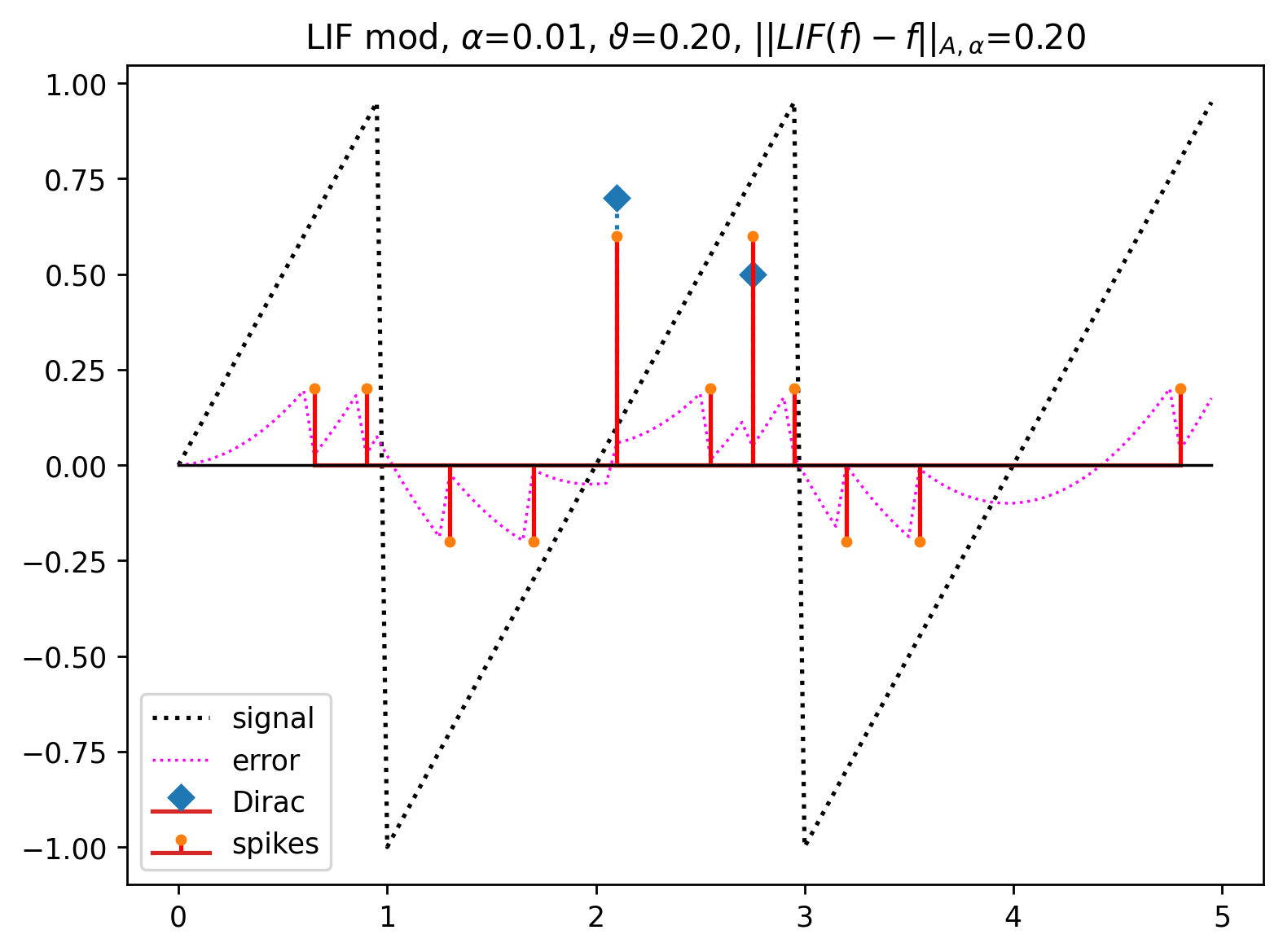}
		\includegraphics[width=0.325\textwidth]{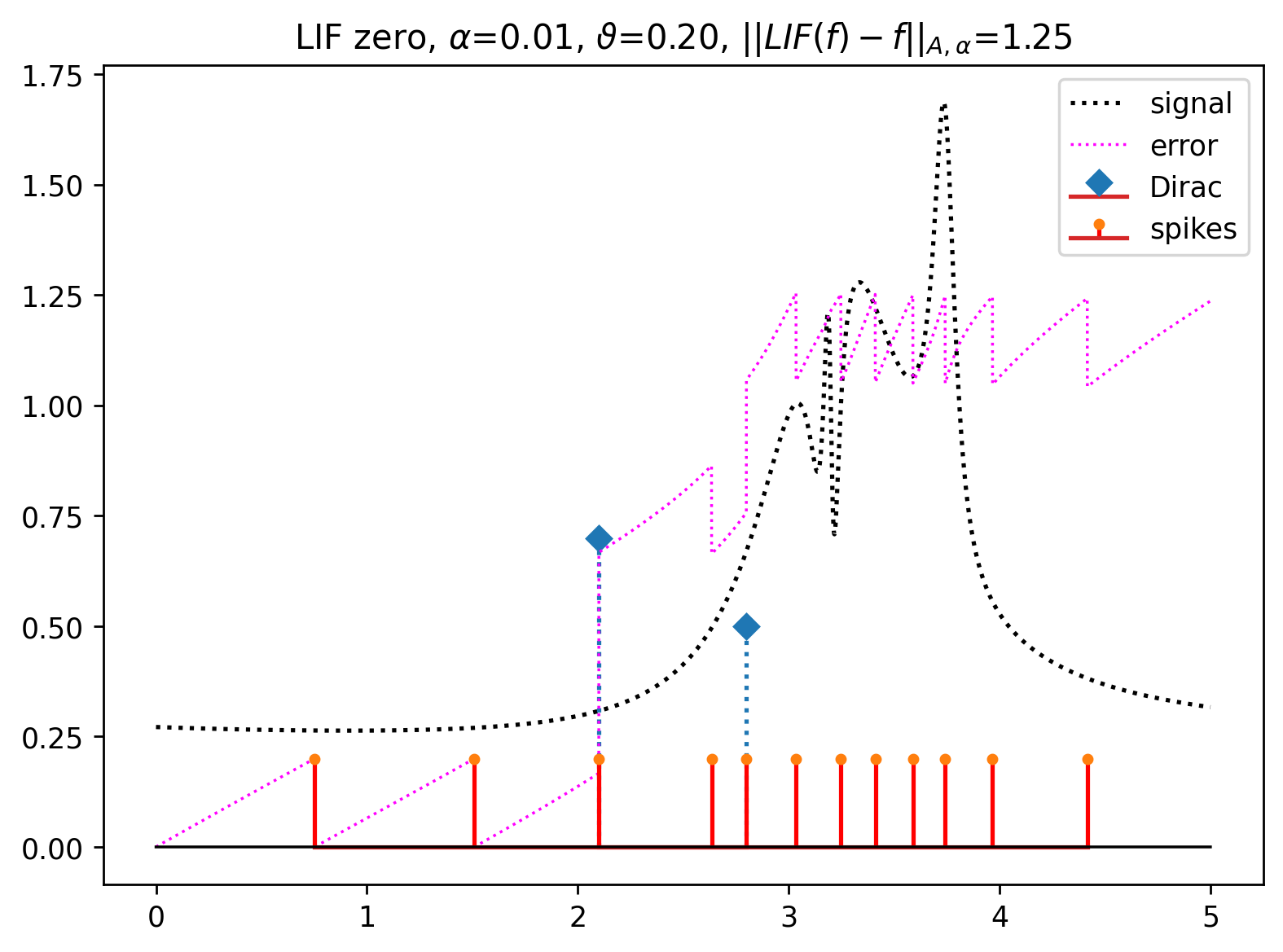}
		\includegraphics[width=0.325\textwidth]{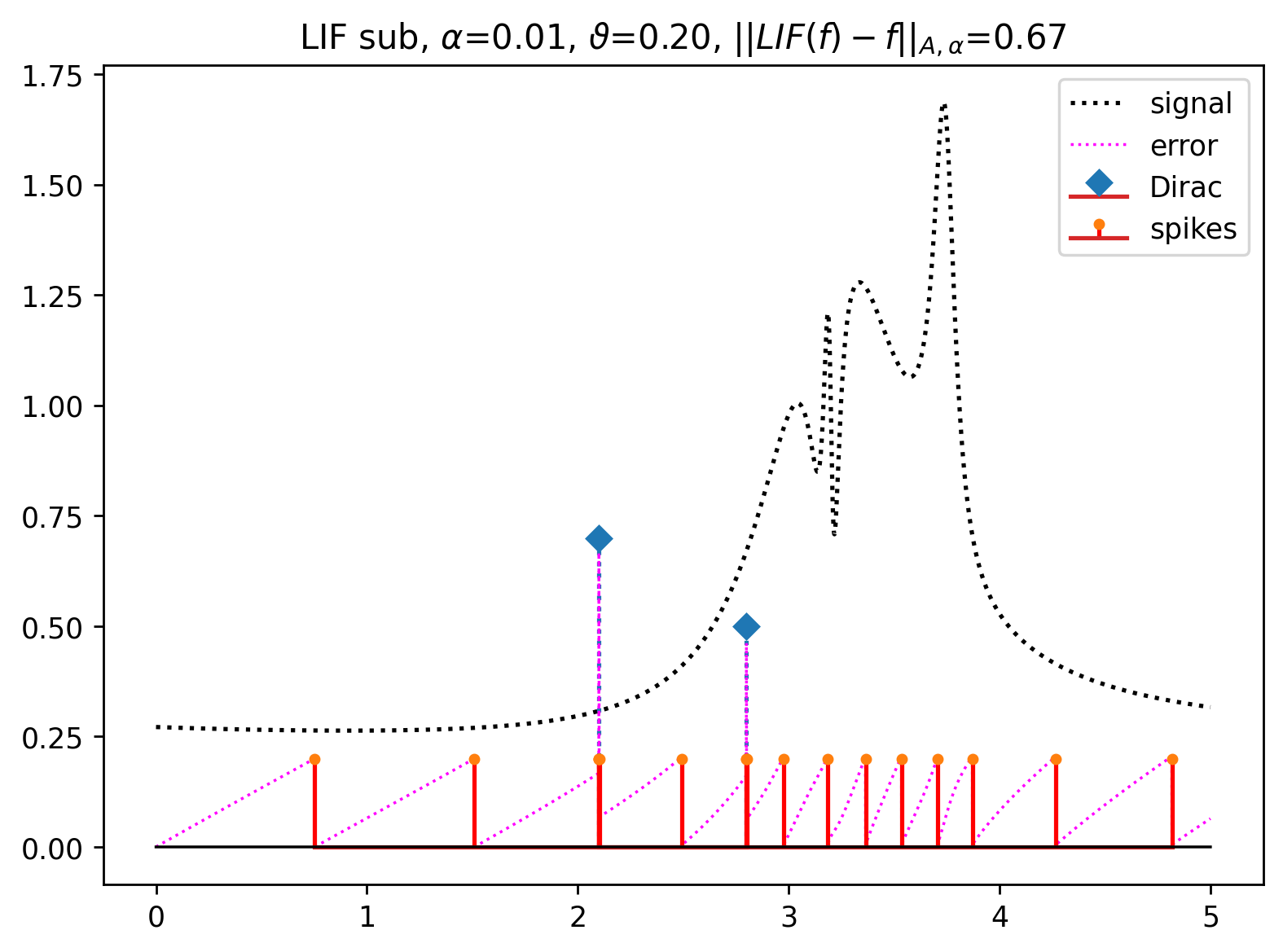}
		\includegraphics[width=0.325\textwidth]{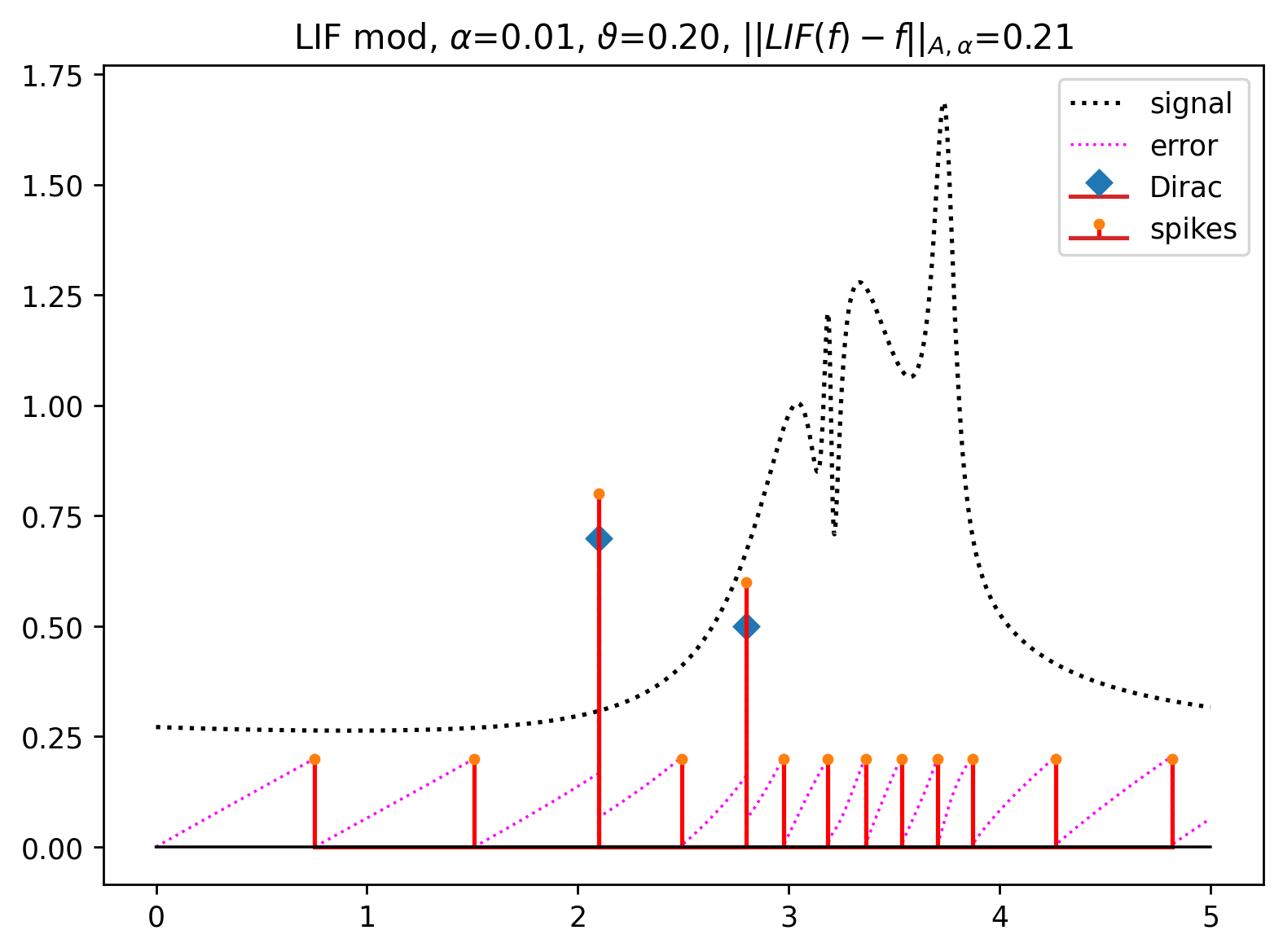}
  		\caption{Examples of Dirac-superimposed signals $f$ (waves (dotted line) with injected Dirac impulses) that are processed by LIF with 
			refractory time $t_r = \mbox{discretization time step}\,\,  \Delta t = 0.01$ showing 
			$\mbox{err}(T):=\int_0^T e^{-\alpha (T - t)}(\mbox{LIF}_{\alpha, \vartheta}(f)(t)-f(t)) dt$  as magenta curve. Due to our theory $\max_t \mbox{err}(t)$ keeps below the threshold for {\it reset-to-mod}; {\it reset-by-subtraction} shows a similar $\mbox{err}$-profile except in the neighborhood of the Dirac impulses while the $\mbox{err}$-profile of {\it reset-to-zero} derails after the first Dirac impulse.
			}
	\label{fig:LIF}
\end{figure}

Injecting weighted Dirac pulses the neuron's potential will show discontinuous jumps, and different reset variants are reasonable from an algorithmic point of view.
Beyond the prevalent variants of {\it reset-to-zero} and {\it reset-by-subtraction}, see e.g.~\cite{snnTorch2021}, 
recently we introduced {\it reset-to-mod} as a third option, see~\cite{moser2023arXiv_SNNAlexTop}. 
{\it reset-to-zero} means that the neuron's potential is reinitialized to zero after firing, while {\it reset-by-subtraction} subtracts the $\vartheta$-potential $u_{\vartheta}$ from the membrane's potential that triggers the firing event. The third variant,  {\it reset-to-mod}, can be understood as 
instantaneously cascaded application of {\it reset-by-subtraction} according to the factor $n$ by which the membrane's potential $u$ exceeds the threshold, i.e.
$u = n \vartheta + r$, $r \in (-\vartheta, \vartheta)$. This means that {\it reset-to-mod} is the limit case of {\it reset-by-subtraction} with the 
refractory time $t_r$ approaching to zero. In this case the residuum $r$ results from a modulo computation and the amplitude of the triggered
spike is set to $n\, \vartheta$, following the understanding the firing-reset mechanism as physically plausible charging-discharging event. 
See Fig.~\ref{fig:LIF} for examples.

\begin{proposition}[Well-Definedness of LIF on Signal Space]
\label{prop:spiketrain}
If $f: [0, \infty)\rightarrow \mathbb{R}$ is locally HK-integrable and bounded almost everywhere with locally finitely many weighted Dirac impulses 
then the LIF process (\ref{eq:LIFsample}) is well defined, i.e., it provides a locally finite sequence $(t_k, s_k)_k$ of spikes and terminates whenever 
restricted to a sub-interval $[a,b] \subset [0, \infty)$. 
\end{proposition}

\begin{proof}
Suppose the contrary that the sequence of sampling points $(t_k)_k$ is infinite, i.e. the sampling process does not terminate in 
$[a,b] \subset [0, \infty)$.
Then there is an accumulation point $t^* \in [a, b]$, such that $|t_{k+1} - t_k| \rightarrow 0$.
As a consequence, we get $|\int_{t_k}^{t_{k+1}} e^{-\alpha (t_{k+1} - t)} f(t) dt| \leq 
|t_{k+1} - t_k| \, \mbox{ess-sup}_{t \in [a,b]} |f(t)| \stackrel{k\rightarrow \infty}{\longrightarrow} 0$,
which
contradicts the spike triggering condition $|\int_{t_k}^{t_{k+1}} e^{-\alpha (t_{k+1} - t)} f(t) dt| \geq \vartheta$.
\end{proof}

Local integrability is a necessary condition to make the LIF integral applicable, and the a.e. boundedness condition is required 
in order to yield a spike train with well separated points in time so that the sampling procedure terminates if applied on a finite time interval. 
Note that $f$ need not be continuous or band-limited. 

\section{Spike Train Quantization under the General Conditions of C1-4}
\label{s:Alex}
First, we recall the discrete version of (\ref{eq:qFormula}) proven in~\cite{moser2024quantization,moser2023arXiv_SNNAlexTop} and formulate it as 
Lemma~\ref{lem:discreteQ}. See Appendix A  for its proof.
\begin{lemma}
\label{lem:discreteQ}
(\ref{eq:qFormula})  holds for any $\hat{f}(t) = \sum_k s_k \delta(t-t_k)$ with locally finitely many time points $t_k>0$ 
if LIF is realized with {\it reset-to-mod}.
\end{lemma}

Together with the idempotence property of projections, i.e.,
 $\mbox{LIF}_{\alpha, \vartheta}\left(\mbox{LIF}_{\alpha, \vartheta}(f)\right) = \mbox{LIF}_{\alpha, \vartheta}(f)$, formula
(\ref{eq:qFormula}) characterizes $\mbox{LIF}_{\alpha, \vartheta}$ with reset-to-mod as quantization operator in the Alexiewicz norm. 
Next we will show that Lemma~\ref{lem:discreteQ} can be generalized in a straightforward way to the much wider class of
signals satisfying C1-4. See Appendix B  for its proof.
\begin{theorem}[LIF with {\it \bf reset-to-mod} as $\|.\|_{A,\alpha}$-Quantization for C1-4 Signals]
\label{th:quantization}
For any signal  $f:[0, \infty) \rightarrow \mathbb{R}$ satisfying the conditions C1-4,  
$\mbox{LIF}_{\vartheta, \alpha}(f)$ is a $\vartheta$-quantization of $f$ w.r.t the geometry induced by the weighted Alexiewicz norm $\|.\|_{A,\alpha}$, 
i.e., the resulting spike amplitudes are multiples of $\vartheta$ with the quantization error is bounded by (\ref{eq:qFormula}), provided that
LIF is realized with {\it reset-to-mod}.
\end{theorem}


\section{Evaluations}
\label{s:Evaluations}
\begin{figure}[ht]
	\centering
	\includegraphics[width=0.325\textwidth]{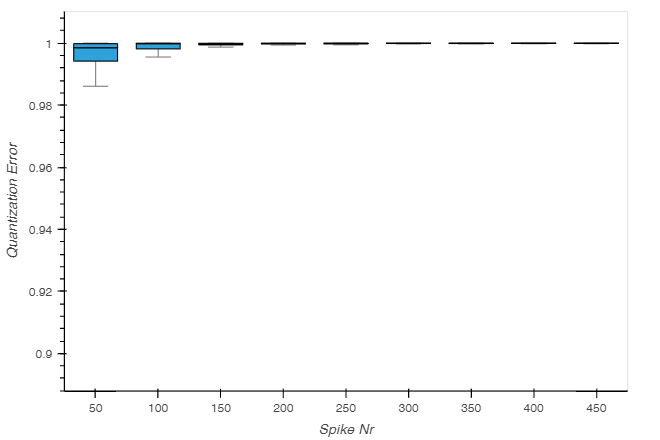}
	\includegraphics[width=0.325\textwidth]{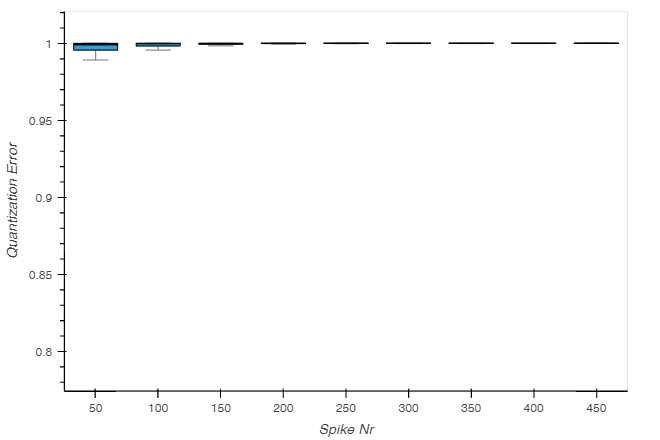}
	\includegraphics[width=0.325\textwidth]{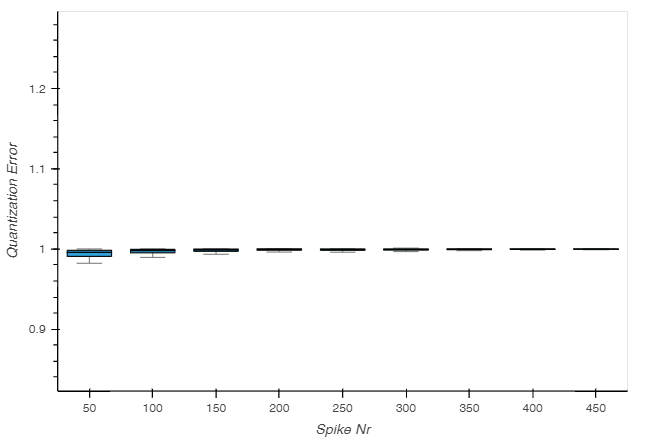}
	\includegraphics[width=0.325\textwidth]{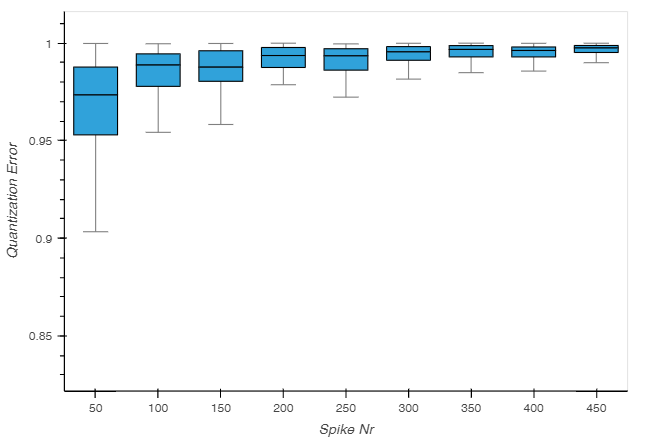}
	\includegraphics[width=0.325\textwidth]{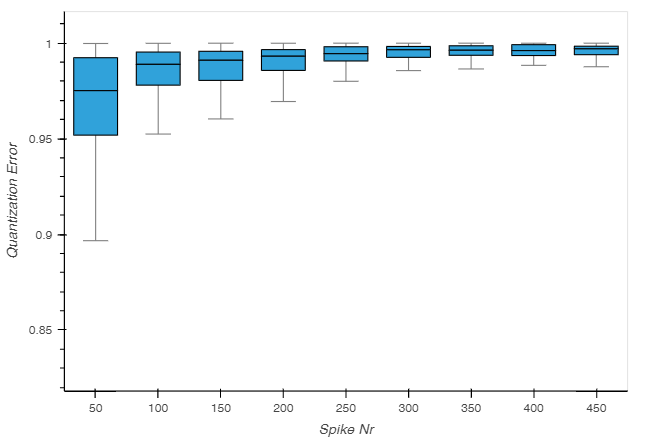}
	\includegraphics[width=0.325\textwidth]{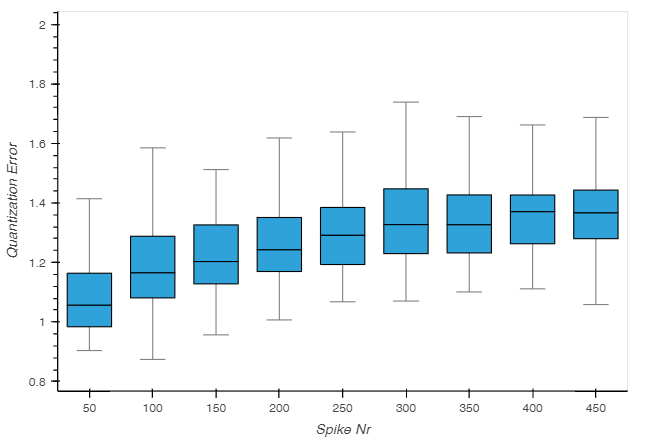}
		\caption{
		Evaluation of (\ref{eq:qFormula}) for {\it reset-to-mod}, {\it reset-by-subtraction} and
		{\it reset-to-zero} (1st/2nd/3rd column), based on spike trains with spike amplitudes in $[-\vartheta, \vartheta]$ and $100$ runs.
		The 1st row refers to $\alpha = 1$ and the 2nd row to $\alpha = 0.1$.		
		}
			\label{fig:QuantizationError1}
\end{figure}

We evaluate the distribution of the quantization error 
(\ref{eq:qFormula}) by means of box whisker diagrams 
for different re-initialization variants depending on the number of weighted Dirac impulses in the input signal and different
distributions of weights. Fig.~\ref{fig:QuantizationError1} shows the result for incoming Dirac weights that are below threshold.
In this case, (\ref{eq:qFormula}) holds for {\it reset-to-mod} and {\it reset-by-subtraction}. 
Under these circumstances, therefore, {\it reset-by-subtraction} and {\it reset-to-mod} can be seen as mutual approximations of each other.
This way, the quantization formula (\ref{eq:qFormula}) is valid or at least approximately valid also for {\it reset-by-subtraction}.
{\it reset-to-zero} only coincides with the others if there are no Dirac impulses and the input signal is bounded and of bounded variation, or in the case of 
Dirac weights below threshold and larger leaky parameters, see first row of Fig.~\ref{fig:QuantizationError1}. In this case the integral $\int_0^T f(t) dt$ changes continuously w.r.t $T$, so that the triggering potential is the same for all reset variants.

\begin{figure}[ht]
	\centering
	\includegraphics[width=0.325\textwidth]{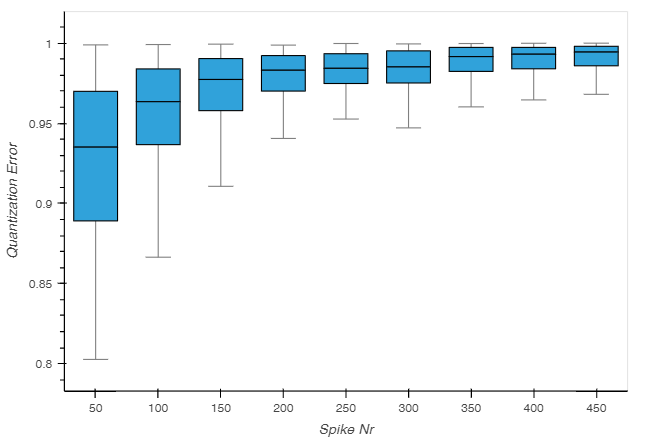}
	\includegraphics[width=0.325\textwidth]{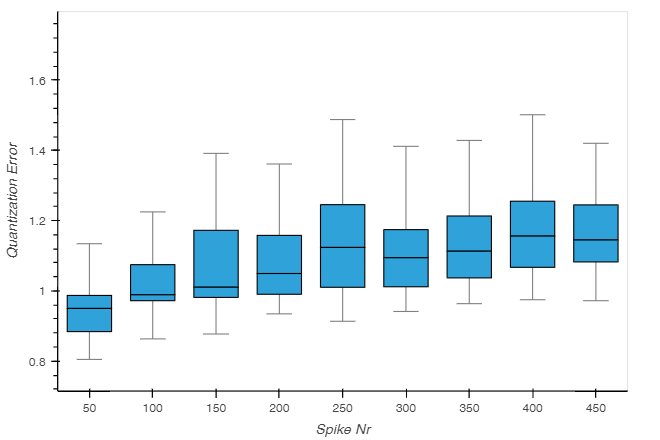}
	\includegraphics[width=0.325\textwidth]{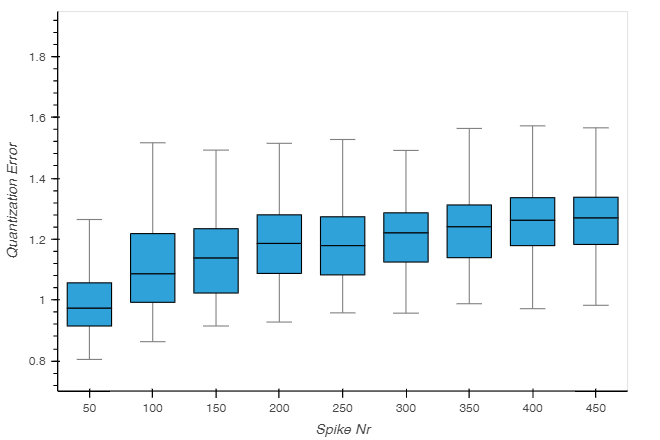}
	\includegraphics[width=0.325\textwidth]{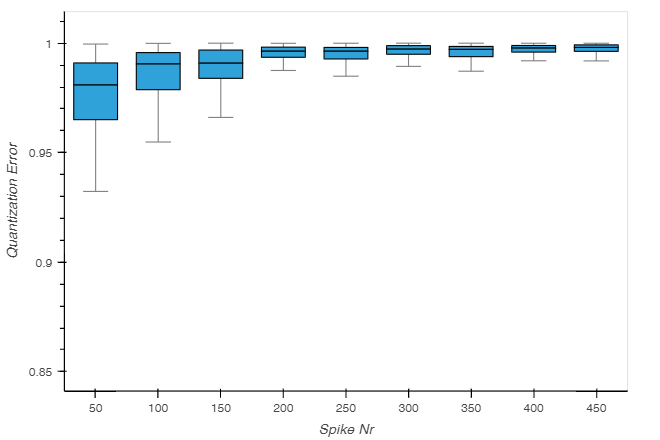}
	\includegraphics[width=0.325\textwidth]{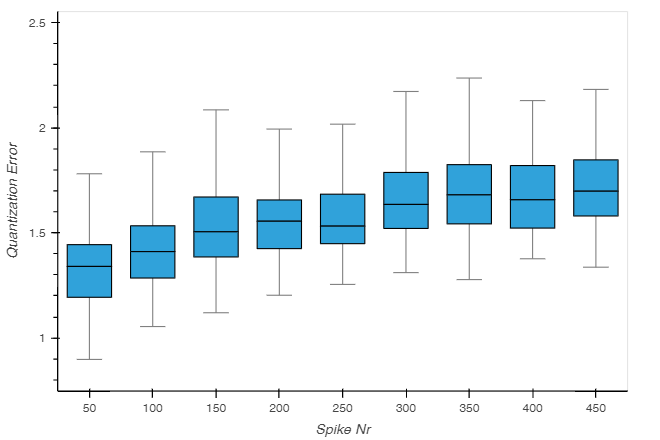}
	\includegraphics[width=0.325\textwidth]{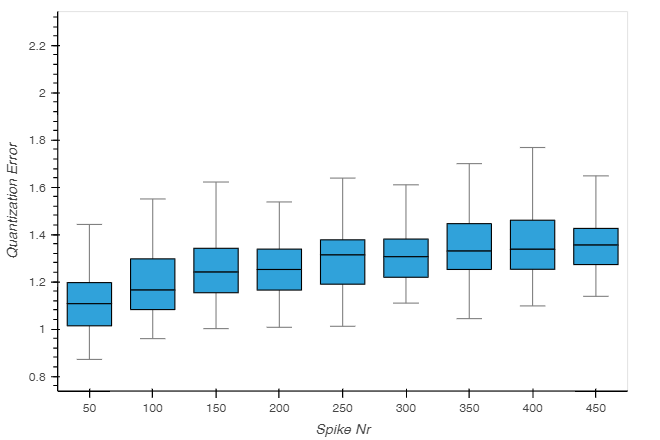}
		\caption{The same as in Fig.~\ref{fig:QuantizationError1} but with spike amplitudes in $[-3/2 \vartheta, 3/2 \vartheta]$.}
			\label{fig:QuantizationError2}
\end{figure}

However, the presence of Dirac impulses has a strong impact, so that with increasing weights of Dirac impulses 
the variants show increasing different behavior. For {\it reset-to-mod} and {\it reset-by-subtraction}, with increasing number of spikes one can observe a concentration of measure effect, see, e.g.,~\cite{Vershynin2018}). This effect is a direct consequence of the quantization formula (\ref{eq:qFormula}) that enforces the distribution of the errors close the surface of the Alexiewicz ball.
However, as shown in Fig.~\ref{fig:QuantizationError2}, if the incoming spike amplitudes are not below threshold anymore Eqn.~(\ref{eq:qFormula}) 
only holds for {\it reset-to-mod} in a strict sense. In this case the measure of concentration effect becomes more apparent for smaller leaky parameters. 
For the other variants the quantization error increases in average with the number of spikes, and the theoretical bound~(\ref{eq:qFormula}), proven for {\it reset-to-mod}, does not hold anymore.

\section{Conclusion}
\label{s:Conclusion}
In this paper we look on the leaky-integrate-and fire (LIF) model as non-linear mapping from some signal space into the space of spike trains.
LIF with reset-by-subtraction with zero refractory time turns out to be an idempotent projection that satisfies the outlined quantization formula based on the Alexiewicz norm. In this limit case the reset mode boils down to a modulo-based scheme, which we refer to as {\it reset-to-mod}. 
The proof is provided for a general class of signals and sheds light on how 
LIF transforms hybrid Dirac-superimposed continuous-time signals into spike trains. The general signal class mentioned goes far beyond what is used in standard signal processing and thus in machine learning-based processing of time-varying signals, which is based on a digitised version of the signal at the input.
The outlined approach is a starting point to link signal processing based on analogue-to-spike converters with spiking neural networks (SNN).
As a byproduct, our analysis gives rise to rethinking the re-initialization modes {\it reset-to-zero} and {\it reset-by-subtraction} that are commonly used in the context of SNN, this way contributing to the mathematical foundation of this field. 
As pointed out by means of simulation experiments the standard re-initialization modes satisfy the quantization formula only under restricted conditions while our proposed variant {\it reset-to-mod} satisfies the derived quantization bound under general conditions.
Examples can be found in the github repository~\url{https://github.com/LunglmayrMoser/AlexSNN}.
The experiments demonstrate in which cases {\it reset-to-mod} can be looked at as reasonable approximation of {\it reset-by-subtraction} and in which not. For a moderate discrete portion in the signal in terms of weighted Dirac impulses in the range of the threshold {\it reset-by-subtraction} and {\it reset-by-mod} can be looked as mutual approximations. 
In the near future, we will use these insights both for signal reconstruction from spike trains with threshold-based error bounds on the signal processing side and rethink backpropagation for SNNs from a machine learning perspective.

\section*{Acknowledgements}
This work was supported (1) by the 'University SAL Labs' initiative of Silicon Austria Labs (SAL) and its Austrian partner universities for applied fundamental research for electronic based systems, (2) by Austrian ministries BMK, BMDW, and the State of Upper-Austria in the frame of SCCH, part of the COMET Programme managed by FFG, and (3) by the {\it NeuroSoC} project funded under the Horizon Europe Grant Agreement number 101070634.

\section*{Appendix}
\subsection*{Appendix A: Proof of Lemma~\ref{lem:discreteQ}, see~\cite{moser2024quantization}}
\label{App:A}

\begin{wrapfigure}{r}{5.5cm}
  \includegraphics[width=5cm]{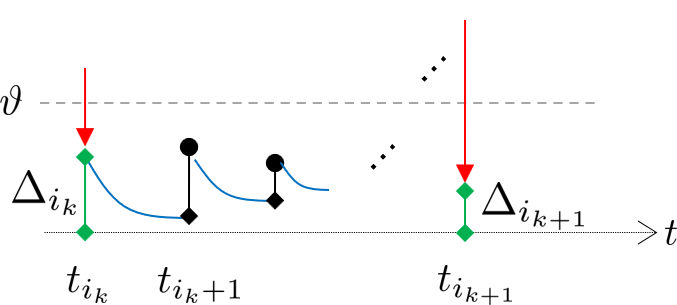}
	\caption{Illustration of Eqn.~(\ref{eq:quantDeltaRecursion}). The red arrows indicate reset by {\it reset-to-mod}.}
 		\label{fig:quantProof}
\end{wrapfigure}

First of all, we introduce the following operation $\oplus$, which is associative and can be handled with like the usual addition if adjacent elements $a_i$ from a spike train $\eta = \sum_i a_i \delta_{t_i}$ are aggregated:
\begin{equation}
\label{eq:pseudoaddition}
a_i \oplus a_{i+1} :=  e^{-\alpha (t_{i+1}- t_{i})} a_i  +  a_{i+1}.
\end{equation}
This way we get a simpler notation when aggregating convolutions, e.g., 
\[
a_i \oplus \ldots \oplus a_j = \sum_{k=i}^j e^{-\alpha (t_{j}- t_{k})} a_k.
\] 

For the discrete version we re-define 
$a_{i_k} \oplus a_{i_{k+1}} :=  \beta^{(i_{k+1}- i_{k})} a_{i_k}  +  a_{i_{k+1}}$, if $i_{k}$ and $i_{k+1}$ refer to adjacent spikes at time $i_k$, resp. $i_{k+1}$. Further, we denote $q[z]:= \mbox{sgn}(z)\, \left\lfloor |z|\right\rfloor$ which is the ordinary quantization due to integer truncation, e.g. $q[1.8] = 1$, $q[-1.8]=-1$, where $\left\lfloor |z| \right\rfloor = \max\{n\in \mathbb{N}_0:\, n \leq |z|\}$.  

After fixing notation let us consider a spike train $\eta = \sum_j a_i \delta_{t_{i}}$. 
Without loss of generality we may assume that $\vartheta = 1$.
We have to show that $\|\mbox{LIF}_{1, \alpha}(\eta)  - \eta\|_{A, \alpha} < 1$, which is equivalent 
to the discrete condition that $\forall n: \max_n \left|\sum_{i=1}^n \hat{a}_i \right| < 1$,
where  $\eta - \mbox{LIF}_{1, \alpha}(\eta) = \sum_i \hat{a}_i \delta_{t_i}$.
Set $\hat{s}_k := \hat{a}_0 \oplus \cdots \oplus \hat{a}_k$. We have to show that
$\max_{k}|\hat{s}_k| < 1$. The proof is based on induction and leads the problem back to the standard quantization by truncation.

Suppose that at time $t_{i_{k-1}}$ after re-initialization by {\it reset-to-mod} we get the residuum $\Delta_{i_{k-1}}$ as membrane potential that is the starting point for the integration after $t_{i_{k-1}}$. 
Note that 
\begin{equation}
\mbox{LIF}_{1, \alpha}(\eta)|_{t = t_k} = q(\Delta_{i_{k-1}} \oplus a_{i_{k-1}+1} \cdots \oplus a_{i_{k}}) \nonumber
\end{equation}
Then, as illustrated in Fig.~\ref{fig:quantProof} the residuum $\Delta_{i_{k}}$ at the next triggering event $t_{i_{k}}$ is obtained by the equation
\begin{equation}
\label{eq:quantDeltaRecursion}
\Delta_{i_{k}} = \Delta_{i_{k-1}} \oplus a_{i_{k-1}+1} \oplus \ldots \oplus  a_{i_k} - q[\Delta_{i_{k-1}} \oplus \ldots \oplus  a_{i_k}].
\end{equation}
Note that due to the thresholding condition of LIF we have
\begin{equation}
\label{eq:thcond}
|\Delta_{i_{k}} \oplus a_{i_{k}+1} \oplus \ldots \oplus  a_j| < 1
\end{equation}
for $j \in \{i_{k}+1, \ldots, i_{k+1}-1\}$.
For the $\oplus$-sums $\hat{s}_{i_k}$ we have
\begin{equation}
\label{eq:ahat}
\hat{s}_{i_{k+1}} = \hat{s}_{i_{k}} \oplus a_{i_{k}+1} \cdots a_{i_{k+1}-1} \oplus 
\left( 
a_{i_{k+1}} - q[\Delta_{i_{k}} \oplus a_{i_{k}+1} \oplus \ldots \oplus a_{i_{k+1}}]
\right).
\end{equation}

Note that $\hat{s}_0 = \Delta_{i_0}= a_0 - q[a_0]$, then for induction we assume that up to index $k$ to have
\begin{equation}
\label{eq:quantInduction}
\hat{s}_{i_k} = \Delta_{i_k}.
\end{equation}

Now, using (\ref{eq:quantInduction}), Equation~(\ref{eq:ahat}) gives
\begin{eqnarray}
\label{eq:induction}
\hat{s}_{i_{k+1}} & = & \Delta_{i_k} \oplus a_{i_{k}+1} \oplus \ldots \oplus 
a_{i_{k+1}-1} \oplus 
\left( 
a_{i_{k+1}} - q[\Delta_{i_{k}} \oplus a_{i_{k+1}} \oplus \ldots \oplus a_{i_{k+1}}]
\right) \nonumber \\
& = & \Delta_{i_k} \oplus a_{i_{k}+1} \oplus \ldots \oplus 
a_{i_{k+1}-1} \oplus 
a_{i_{k+1}} - q[\Delta_{i_{k}} \oplus a_{i_{k+1}} \oplus \ldots \oplus a_{i_{k+1}}], \nonumber\\
 & = & \Delta_{i_{k+1}}
\end{eqnarray}
proving (\ref{eq:quantInduction}), which together with (\ref{eq:thcond}) ends the proof showing that 
$|\hat{s}_k|<1$ for all $k$.

\subsection*{Appendix B: Proof of Theorem~\ref{th:quantization}}
\label{App:B}
Given $f$ satisfying C1-4. Applying LIF with reset-to-mod we get 
$s(.) := \sum_k s_k \delta(.- t_k) := \mbox{LIF}_{\vartheta, \alpha}(f) $ with spike amplitudes $s_k \in \vartheta \mathbb{Z}$ and time points $t_k$.
Because of C4 we have $s_0 = 0$.
Let us consider the approximation $\hat{f}: [0, \infty) \rightarrow \mathbb{R}$, $\hat{f}(t):= \sum_k a_k \delta(.- t_k)$ given by
$a_{k+1} := \int_{(t_k, t_{k+1}]} e^{-\alpha (t_{k+1} - t)} f(t) dt $ and $a_0 = 0$.
Note that by construction we get 
\begin{equation}
\label{eq:fhatf}
\mbox{LIF}_{\vartheta, \alpha}(\hat{f}) = \mbox{LIF}_{\vartheta, \alpha}(f),
\end{equation}
\begin{equation}
\label{eq:Thatf}
T \in (t_k)_k \Rightarrow \int_0^T e^{-\alpha (T - t)} f(t) dt = \int_0^T e^{-\alpha (T - t)} \hat{f}(t) dt, 
\end{equation}
and
\begin{equation}
\label{eq:indStep}
\int_{(t_k, T]} e^{-\alpha (T - t)} f(t) dt = 
\left\{
\begin{array}{lcl}
0                 & & \mbox{if}\,\, T \in (t_k, t_{k+1}) \\
a_{k+1} - s_{k+1} & & \mbox{if}\,\, T = t_{k+1} 
\end{array}
\right.
.
\end{equation}
Equation~(\ref{eq:indStep}) implies
\begin{eqnarray}
\label{eq:indSup}
& & \sup_{T \in (t_n, t_{n+1}]}\left|\int_0^T  e^{-\alpha (T - t)} (\hat{f}(t) - s(t))dt\right| 
= \max \left\{\left|\int_{[0, t_n]}  e^{-\alpha (t_n - t)} (\hat{f}(t) - s(t))dt\right|, \right. \nonumber \\
& & 
\left. \left|\int_{[0, t_n]}  e^{-\alpha (t_n - t)} (\hat{f}(t) - s(t))dt 
+  1_{\{T=t_{n+1}\}}(T) \int_{(t_n, t_{n+1}]} e^{-\alpha (T - t)} (\hat{f}(t) - s(t))dt\right|
\right\},\nonumber
\end{eqnarray}
where $1_I(t)=1$ if $t\in I$ and $0$ else, 
which together with C4, i.e., $\int_{[0, t_1)} e^{-\alpha (t_n - t)} (\hat{f}(t) - s(t))dt = 0$, proves the induction step for showing
\begin{equation}
\label{eq:indSupremun}
\sup_{T \geq 0}\left|\int_0^T  e^{-\alpha (T - t)} (\hat{f}(t) - s(t))dt\right| = 
\sup_{T \in (t_k)_k}\left|\int_0^T  e^{-\alpha (T - t)} (\hat{f}(t) - s(t))dt\right|.
\end{equation}
The proof ends by putting all together 
\begin{eqnarray}
\|f - \underbrace{\mbox{LIF}_{\vartheta, \alpha}(f)}_{s}\|_{A, \alpha} & =^{(\ref{eq: AlexNorm})} &
\sup_{T\geq 0} \left|\int_{0}^T e^{-\alpha (T - t)} (f(t) - s(t))dt  \right|  \nonumber \\
&=^{(\ref{eq:LIFsample})}& \sup_{k\geq 0} \left|\int_{0}^{t_k} e^{-\alpha (T - t)} (f(t) - s(t))dt  \right| \nonumber \\
& =^{(\ref{eq:Thatf})}  & 
\sup_{k\geq 0} \left|\int_{0}^{t_k} e^{-\alpha (T - t)} (\hat{f}(t) - s(t))dt  \right| \nonumber \\
& =^{(\ref{eq:indStep})} &
\sup_{T\geq 0} \left|\int_{0}^T e^{-\alpha (T - t)} (\hat{f}(t) - s(t))dt  \right| \nonumber \\
& =^{(\ref{eq:fhatf})} & 
\|\hat{f} - \mbox{LIF}_{\vartheta, \alpha}(\hat{f})\|_{A, \alpha}    \nonumber \\
& <& \vartheta. \nonumber
\end{eqnarray}


\bibliographystyle{unsrtnat}
\bibliography{references}
\end{document}